# Heuristic-Informed Mixture of Experts for Link Prediction in Multilayer Networks

**Lucio La Cava**, **Domenico Mandaglio**, **Lorenzo Zangari**, **Andrea Tagarelli**

Dept. DIMES, University of Calabria, 87036 Rende (CS), Italy

{lucio.lacava, d.mandaglio, lorenzo.zangari, tagarelli}@dimes.unical.it

## Abstract

Link prediction algorithms for multilayer networks are in principle required to effectively account for the entire layered structure while capturing the unique contexts offered by each layer. However, many existing approaches excel at predicting specific links in certain layers but struggle with others, as they fail to effectively leverage the diverse information encoded across different network layers. In this paper, we present MoE-ML-LP, the first Mixture-of-Experts (MoE) framework specifically designed for multilayer link prediction. Building on top of multilayer heuristics for link prediction, MoE-ML-LP synthesizes the decisions taken by diverse experts, resulting in significantly enhanced predictive capabilities. Our extensive experimental evaluation on real-world and synthetic networks demonstrates that MoE-ML-LP consistently outperforms several baselines and competing methods, achieving remarkable improvements of +60% in Mean Reciprocal Rank, +82% in Hits@1, +55% in Hits@5, and +41% in Hits@10. Furthermore, MoE-ML-LP features a modular architecture that enables the seamless integration of newly developed experts without necessitating the re-training of the entire framework, fostering efficiency and scalability to new experts, paving the way for future advancements in link prediction.

## 1 Introduction

Several real-world systems with complex dependency structures, such as social and biological networks, can profitably be modeled through the formalism of *multilayer networks* [Kivelä *et al.*, 2014], in which two or more graphs (aka *layers*) are interconnected and represent different types or contexts of relationships between the entities of a system. Like in simple (i.e., single-layer) networks, one fundamental problem in multilayer networks is *link prediction*, which is to estimate the likelihood that a link exists between two nodes in one layer, based on both intra- and inter-layer information. In recent years, several works have addressed this problem using various strategies derived from approaches that are shown to be effective in the single-layer case. In this regard, link prediction research has often focused on *heuristic methods* (e.g., Common Neighbors, Adamic-Adar, etc.), which estimate the likelihood of a link between two nodes using topologically simple, hand-crafted metrics based on node and edge properties. Despite their simplicity, heuristic methods serve as strong baselines for link prediction, as they can effectively *inform pairwise structural information* specific to node pairs, making them useful for predicting linkage. Given their recognized usefulness for link prediction, heuristics have also been incorporated into complex learning methods based on Graph Neural Networks (GNNs), such as in [Zhang and Chen, 2018; Yun *et al.*, 2021; Chamberlain *et al.*, 2023; Wang *et al.*, 2024]. Although GNNs have achieved strong performance in tasks like node and graph classification, their node-based representation learning design limits their ability to capture link-specific information [Zhang *et al.*, 2021]. Consequently, in some cases, classic heuristics might achieve performance comparable to GNNs in the link prediction task.

However, both traditional heuristic approaches and GNN-based methods typically adopt a one-size-fits-all strategy, uniformly applying the same approach to all target node pairs. This approach has limitations, since different node pairs often require distinct heuristics for accurate predictions [Ma *et al.*, 2024]. This issue is even more pronounced in multilayer networks, where the same pair of nodes may exhibit different structural properties across layers.

In the last few years, with the advancement of Large Language Models (LLMs), the paradigm of *Mixture-of-Experts (MoE)* [Jacobs *et al.*, 1991] has gained renewed attention. MoE is a well-established machine learning method based on the divide-and-conquer principle: the outputs of different parts of a model, called *experts*, are combined and supervised by a gating or routing network [Masoudnia and Ebrahimpour, 2014] that decides which experts to rely on more for a given input. Recently, this paradigm has also been explored in addressing graph representation learning problems on simple graphs, such as node/graph classification and link prediction, yielding promising results [Hu *et al.*, 2022; Ma *et al.*, 2024]. Nevertheless, in the context of multilayer networks, the benefits of the MoE paradigm for link prediction have yet to be investigated. We believe that addressing multilayer link prediction based on a MoE approach can not only lead to boost the performance of individual ex-perts, by reducing their algorithmic biases, but also to mitigate inherent challenges arising in different link prediction approaches for multilayer networks, such as the integration of information from all layers by harnessing always the same heuristic strategy [Aleta *et al.*, 2020; Zangari *et al.*, 2024], which can be detrimental for link prediction [Yun *et al.*, 2021; Wang *et al.*, 2024; Masoudnia and Ebrahimpour, 2014].

To fill this gap in the literature, we propose MoE-ML-LP, a heuristic-informed mixture-of-experts framework for link prediction in multilayer networks. A key novelty of MoE-ML-LP is the use of heuristics that incorporate essential topological criteria for link prediction, which are feed into a gating network to guide the activation of expert models for multilayer link prediction tasks. Our designed MoE strategy has a two-fold effect. First, the heuristic-driven selection of multilayer experts ensures that the framework captures essential structural information for link prediction. Second, by utilizing an arbitrary ensemble of specialized multilayer experts along with a gating mechanism to select the most suitable expert for each node-pair in each layer, MoE-ML-LP maintains both modularity and expert-level performance.

**Contributions** of this work are summarized as follows:

• We propose MoE-ML-LP, a novel MoE framework specifically designed for link prediction in multilayer networks. MoE-ML-LP leverages multilayer-enhanced heuristics to combine individual predictions of expert networks into an enhanced final prediction. To the best of our knowledge, this is the first approach that integrates heuristics within a MoE architecture for link prediction in multilayer networks.

• The MoE architecture enables MoE-ML-LP to gain a complementary perspective over the intricate across-layer dependencies compared to traditional methods, leading to unprecedented performance gains over both traditional multilayer baselines and the most competitive individual approaches. MoE-ML-LP sets a new benchmark in multilayer link prediction on real-world networks, delivering an average improvement of +60% in MRR, +81.9% in Hits@1, +54.5% in Hits@5, and +40.6% in Hits@10, thereby demonstrating remarkable predictive power.

• MoE-ML-LP is designed with a focus on modularity and efficiency. It preserves and exploits the unique strengths and specialized approaches of individual experts, supporting diverse prediction strategies. Furthermore, its modular architecture allows the seamless integration of newly released experts without the need to re-train the entire framework, resulting in a particularly efficient approach.

## 2 Related Work

**Link prediction on multilayer networks.** Link prediction on simple graphs has traditionally been tackled using heuristic approaches relying on topological node similarity measures [Zhang, 2022], such as Common Neighbors [Liben-Nowell and Kleinberg, 2003], Jaccard score, Adamic-Adar [Adamic and Adar, 2003], and personalized PageRank [Brin and Page, 1998]. Given the growing attention garnered by multilayer networks, researchers have been prompted to generalize these heuristics to the multilayer case, where the main challenge is to generate link scores that account for the interplay among different layers. For instance, [Rossetti *et al.*, 2011] defines various neighborhood notions in multilayer networks, which are key to building multilayer versions of the above mentioned heuristics. Similarly, [Tillman *et al.*, 2020] propose various multilayer heuristics that leverage both single-layer methods and correlation measures across different layers. However, these methods require the manual selection of the heuristic to be applied to each network, while in general, different heuristics could be effective for different layers and/or node-pairs [Yun *et al.*, 2021; Wang *et al.*, 2024; Zangari *et al.*, 2024].

Graph Neural Networks have been employed to enhance link prediction beyond traditional heuristics. Initially successful on simple graphs, these techniques have also been extended to multilayer networks. GATNE [Cen *et al.*, 2019] and BPHGNN [Fu *et al.*, 2023] are designed for attributed multiplex heterogeneous networks. The former learns node embeddings at each layer by combining a base embedding shared by each entity and an edge embedding. The latter uses a contrastive learning strategy to learn node representations from both a local and a global perspective of the multilayer representation. DMGI [Park *et al.*, 2020] and HDMI [Jing *et al.*, 2021] learn node representations in attributed multiplex networks following the Deep Graph Infomax method [Velickovic *et al.*, 2019]. [Zangari *et al.*, 2024] combine different types of overlapping multilayer neighborhoods through an attention mechanism to predict links at each layer. MAGMA [Coscia *et al.*, 2022] generates graph association rules by identifying all frequent patterns in a network via multiplex graph mining, then it assigns a score to each node-pair by finding the occurrences of each rule across the network. Although methods like GATNE and BPHGNN are designed for heterogeneous networks, and MAGMA employs a different paradigm compared to them, we emphasize that they can be seamlessly exploited by MoE-ML-LP, since it is independent of the architectural design of the experts.

**Mixture of Experts.** Mixture of Experts (MoE) is an efficient paradigm for integrating the capabilities of specialized models known as *experts* through gating or routing networks. Originally introduced in [Jacobs *et al.*, 1991] as a divide-and-conquer strategy for breaking down complex tasks into smaller sub-tasks, handled by simpler models, MoE has gained particular attention in recent years due to its applicability to novel problems [Masoudnia and Ebrahimpour, 2014; Cai *et al.*, 2024]. MoE has demonstrated significant effectiveness in both Natural Language Processing [Shazeer *et al.*, 2017; Zhou *et al.*, 2022; Artetxe *et al.*, 2022; Du *et al.*, 2022; Jiang *et al.*, 2024] and Computer Vision [Riquelme *et al.*, 2021], enhancing efficiency and performance by exploiting the inherent clustered structure of the data [Chen *et al.*, 2022].

The link prediction problem has also encountered the MoE paradigm. In [Ma *et al.*, 2024], the problem is addressed by leveraging a mixture of structural and exogenous feature information, although it is limited to single-layer graphs.

Prompted by its promising results in the single-layer case, we design *the first MoE-based approach for link prediction in multilayer networks*, which poses unique challenges due to diverse connectivity patterns across layers. Indeed, the heuristic information related to a target edge on a specific layer should also account for information from other layers, in order to optimally rely on the most suitable experts through a holistic view of the multilayer network. Tackling these challenges requires a modular, fine-grained approach that leverages inter-layer information and automatically selects the most suitable experts for each target node-pair. By adopting a heuristic-informed MoE paradigm for link prediction in multilayer networks, we dynamically activate the most suitable experts for each target node-pair based on the relevant multilayer structural information surrounding that pair.

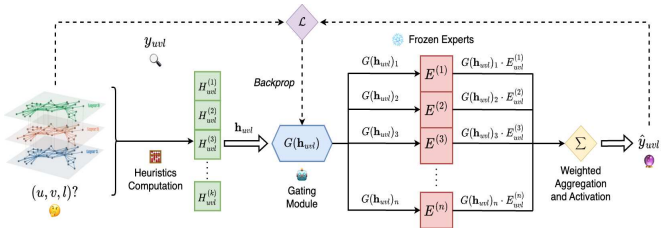

Figure 1: Overview of MoE-ML-LP.

## 3 Preliminary Definitions

**Multilayer networks.** Let $\mathcal{V}$ be a set of *entities* and a $\mathcal{L} = \{1, \ldots, \ell\}$ be a set of *layers* (layer ids), with $|\mathcal{L}| = \ell \geq 2$. We denote a multilayer network as $\mathcal{G} = \langle V_\mathcal{L}, E_\mathcal{L}, \mathcal{V}, \mathcal{L} \rangle$, where $V_\mathcal{L} \subseteq \mathcal{V} \times \mathcal{L}$ is the set of all entity occurrences, or *nodes*, in $\mathcal{L}$, and in particular, $V_l$ is the set of nodes in layer $l$ ($l \in \mathcal{L}$); in the following, we will refer to $\langle v, l \rangle \in V_l$ simply as $v \in V_l$ to denote a node (i.e., occurrence of entity) $v$ in layer $l$. $E_\mathcal{L}$ is the set of edges between nodes belonging to the same layer. Each entity has a node in at least one layer, implying $\mathcal{V} = \bigcup_{l=1..\ell} V_l$, and *inter-layer edges* exist between each node in a layer and its counterpart in a different layer. Also, entities may be associated with external information or attributes. Note that we assume independence from any relation order between the layers; however, if such information is available, it can be exploited.

**Problem statement (Multilayer Link Prediction).** Given a multilayer network $\mathcal{G}$, the multilayer link prediction problem is to estimate the probability of existence of edges in an arbitrary set of layers. More specifically, *for each* layer $l \in \mathcal{L}$, the goal is to learn a function $s : V_l \times V_l \mapsto [0, 1]$ expressing the likelihood of linkage between any pair of nodes in $l$, based on intra- and inter-layer connectivity information.

We emphasize that the problem we consider focuses on performing link prediction across all layers of the network simultaneously, i.e., treating all layers in $\mathcal{L}$ as target layers.

**Heuristics and Experts.** In this work we address the above stated problem based on a machine-learning framework whose core includes *heuristics* as well as *experts*. We distinguish the two types of methods as follows: heuristics are regarded as methods that apply a scoring function based on local/global properties of the input (multilayer) network topology; however, unlike experts, they do not involve optimization of any objective function, nor a training process or parameters to be learned from the data.

## 4 The MoE-ML-LP Framework

To address the multilayer link prediction problem, we propose MoE-ML-LP, a Mixture-of-Experts (MoE) framework. Its main components and data flow are illustrated in Figure 1.

**Overview.** MoE-ML-LP is conceived to be trained on triples $(u, v, l)$ (with $u, v \in V_l$ and $l \in \mathcal{L}$) and associated labels indicating whether a link between $u$ and $v$ exists in layer $l$. The MoE framework consists of three core components: (i) the *multilayer heuristic module*, which is responsible for computing multilayer heuristic scores for a given triple $(u, v, l)$, (ii) the *gating module*, which is designed to learn gating weights that indicate the relevance of each expert to the prediction of a specific triple, and (iii) the *aggregation and activation module*, which leverages these weights to selectively activate experts and aggregate their predictions into a final output.

MoE-ML-LP uses a *heuristic-informed* Mixture-of-Experts framework, central to its effectiveness in multilayer link prediction. Indeed, different heuristics capture (partly) different structural patterns, providing alternative contexts for link prediction. For instance, if certain heuristic scores indicate strong local connections, e.g., a high fraction of common neighbors, the gating module can learn to prioritize experts that excel in shaping dense local structures. Conversely, global heuristics, such as random-walk-based scores, guide the gating module to exploit experts specialized at long-range or cross-layer dependencies.

This heuristic-informed activation strategy hence allows MoE-ML-LP to dynamically adapt to the unique features of each input triple and network. Indeed, by learning how to properly associate specific heuristic scores with the most appropriate experts, MoE-ML-LP optimizes its decision-making process and boosts the prediction capabilities by combining the strengths of different experts into a single probability indicating the likelihood of linkage between $(u, v)$ in layer $l$. A description of the main components of MoE-ML-LP is reported next.

**Multilayer Link Prediction Heuristics.** We consider traditional heuristic-based link prediction methods, originally designed for single-layer graphs (e.g., Common Neighbors, Adamic-Adar), and extend them to a multilayer network setting. Given a heuristic $h$, we denote with $h(u, v, l)$ the linkage score computed by $h$ for a node pair $(u, v)$ in layer $l$. We then define an *aggregated multilayer score* $H(u, v, l)$ that incorporates information from both the target layer $l$ and the other layers in the network:

$$H(u, v, l) = \alpha \cdot h(u, v, l) + \left(1 - \alpha\right) \frac{1}{|\mathcal{L}| - 1} \sum_{l' \in \mathcal{L} \setminus l} h(u, v, l'), \quad (1)$$

where $\alpha \in (0, 1)$ is a smoothing coefficient controlling the contribution of the target layer $l$, while the second term averages the heuristic scores from all other layers $l' \neq l$. Equation (1) allows for a flexible balance between the information from the target layer and the broader multilayer structure when computing the final link prediction score. It should also be noted that our approach is deliberately general to be applied to any single-layer heuristic, while discarding algorithmic biases of the involved heuristic.

Details on the heuristics considered in the experimental evaluation of this study can be found in Appendix A.

**Gating for Mixture of Experts.** The gating module is aimed to determine how much of the particular signals from the heuristics should be routed through the various experts.

For a target triple $(u, v, l)$, let $H_{uvl}^{(i)}$ represent the aggregated multilayer score from the $i$-th heuristic for $(u, v, l)$ (cf. Equation (1)), and $\mathbf{h}_{uvl} \in \mathbb{R}^k$ denote the concatenation the $k$ multilayer heuristics' scores, i.e., $\mathbf{h}_{uvl} = [H_{uvl}^{(1)}, \ldots, H_{uvl}^{(k)}]$.

We are also given a set of $n$ experts $\{E^{(1)}, \ldots, E^{(n)}\}$ for the multilayer link prediction task. Each of these experts processes the triple independently, producing individual predictions on the linkage from $u$ to $v$ in layer $l$, hereinafter denoted as $E_{uvl}^{(i)}$, with $i = 1..n$. Like for the involvement of the heuristics, MoE-ML-LP is agnostic to the individual experts. This also includes incorporating available side-information (i.e., node features) in the link prediction heuristics or experts.

The gating module, we denote with $G(\cdot)$, is defined to receive the vector $\mathbf{h}_{uvl}$ and to output a set of probability weights over the experts. This can be implemented using any artificial neural network and follow different routing strategies [Hazimeh *et al.*, 2021; Clark *et al.*, 2022; Zhou *et al.*, 2022]; in this work, we compute the softmax over the logits yielded by the last linear layer of the gating module, which has been proven simple yet effective [Jordan and Jacobs, 1994; Shazeer *et al.*, 2017; Jiang *et al.*, 2024]. Formally:

$$G(\mathbf{h}_{uvl}) = softmax(g(\mathbf{h}_{uvl})), \quad (2)$$

where $g$ is a multi-layer perceptron that processes the $k$ heuristics' scores to output raw logits, which are subsequently transformed through the softmax activation function into $n$ probabilities representing weights to assign to the experts.

By aggregating the experts' predictions according to the weights assigned by the gating module, a final prediction probability is computed as follows:

$$\hat{y}_{uvl} = \sigma \left( \sum_{i=1}^{n} G(\mathbf{h}_{uvl})_i \cdot E_{uvl}^{(i)} \right) \quad (3)$$

where $G(\mathbf{h}_{uvl})_i$ is the $i$-th component of the $n$-dimensional output of the gating module, and $\sigma$ is the sigmoid function.

**Dense-to-Sparse MoE.** We emphasize that our reference instantiation of MoE-ML-LP utilizes a dense MoE framework, which is feasible due to (i) the manageable number of experts involved, (ii) the choice of keeping them frozen (see later in this section), and (iii) the gating module's "soft routing" which learns *soft* constraints that occasionally activate only a subset of experts, when some are deemed redundant.

Nonetheless, MoE-ML-LP also supports by design a sparse MoE implementation with "hard routing", activating a fixed number of experts during inference by modifying Eq. 2 as follows [Shazeer *et al.*, 2017; Jiang *et al.*, 2024]:

$$G(\mathbf{h}_{uvl}) = softmax(TopK(g(\mathbf{h}_{uvl}))), \quad (4)$$

where $TopK(\cdot)$ denotes a function to retain the original values of the top-$K$ logits $\in \mathbf{R}^n$ produced by $g(\mathbf{h}_{uvl})$, while setting the remaining ones to $-\infty$, thus distributing the routing probability among $K$ models only.

**Freezing experts.** A key feature of MoE-ML-LP is keeping individual experts frozen during training. This strategy not only enhances efficiency but also promotes modularity, contributing to both usability and performance in several ways:

• *Preserving specialization:* Keeping experts frozen preserves their unique capabilities and optimal hyperparameters, which result from different training procedures and optimizations. This ensures their complementary skills, thus avoiding the risk of "flattening" effect of end-to-end training, where the distinct contributions of each expert could be homogenized.

• *Supporting diverse prediction strategies:* Freezing the experts retains their distinct approaches, including the ability to incorporate additional features, such as node attributes or other data sources. This diversity enhances the richness of predictions, as each expert can bring in a unique perspective or leverage different information.

• *Facilitating modularity and efficiency:* Freezing experts ensures modularity, allowing seamless integration of new experts without retraining existing ones. Unlike end-to-end methods, our approach trains only the new expert and gating module, making MoE-ML-LP scalable to new experts. We discuss the **time complexity** of MoE-ML-LP in Appendix C.

• *Preventing imbalanced utilization:* The freezing choice mitigates the risk of "degenerated" behaviors where certain experts consistently receive more consideration leading to better training whereas others remain underutilized or poorly trained [Shazeer *et al.*, 2017].

**Optimization.** We frame the multilayer link prediction task as a binary classification problem, where existing edges in the input multilayer network serve as positive examples, and a set of randomly sampled non-linked node-pairs act as negative examples (cf. Experimental setting).

For any pair of nodes $u, v$ in layer $l$, let $y_{uvl}$ represent the ground-truth label, s.t. $y_{uvl} = 1$ if nodes $u$ and $v$ are linked with each other in layer $l$, and 0 otherwise. We aim to optimize the binary cross-entropy of each predicted link probability made by MoE-ML-LP for each triple $(u, v, l)$. The loss function associated with the overall link-existence scores is:

$$\mathcal{L} = -\sum_{l \in \mathcal{L}} \sum_{(u,v) \in E_l^{(train)}} y_{uvl} \log \left( \hat{y}_{uvl} \right) + \left( 1 - y_{uvl} \right) \log \left( 1 - \hat{y}_{uvl} \right)$$
(5)

where $E_l^{(train)}$ is the set of (positive and negative) training node-pairs in $l$.

## 5 Experimental Evaluation

**Data.** To experiment with MoE-ML-LP, we used publicly available real-world multilayer networks, namely *CsAarhus* [Magnani *et al.*, 2013], *CKM* [Coleman *et al.*, 1957], *Lazega* [Kraatz *et al.*, 2003], *Elegans* [Chen *et al.*, 2006], *DkPol* [Magnani *et al.*, 2022], and *ArXiv* [De Domenico *et al.*, 2015a]. These vary in scope, structural characteristics, and semantics associated with the different layers, thus allowing for a proper validation of our proposed approach. For the sake of brevity, we refer the reader to Appendix B for a detailed description of these networks. In addition, we built synthetic networks of varying sizes based on the Watts-Strogatz generative model. More details can be found in Appendix D.4.

**Assessment Criteria and Goals.** Following prior works on link prediction [Hu *et al.*, 2020], we resort to the Mean Reciprocal Rank (MRR) and Hits@$k$ measures to evaluate MoE-ML-LP, as the primary goal in this task is to rank positive pairs higher than negative ones. Given a query set $Q$, i.e., a set of triples $(u, v, l)$ for which methods are tasked to predict whether the link exists, MRR is the averaged reciprocal rank of the first correctly predicted link across $Q$: $MRR = \frac{1}{|Q|}\sum_{i=1}^{|Q|} \frac{1}{r_i}$, where $r_i$ is the rank position of the first relevant (i.e., positive) link for the $i$-th query. Hits@$k$ is the fraction of queries where at least one relevant link appears in the top-$k$ positions: $Hits@k = \frac{1}{|Q|}\sum_{i=1}^{|Q|} \mathbb{I}(r_i \leq k)$. For both criteria, higher values indicate better performance.

**Experimental setting.** For any input multilayer network, we are interested in evaluating the link prediction outcomes *on all its layers*, at the same time. Note that our setting is *transductive*, i.e., all nodes are available at training time.

We split the set of edges of a multilayer network into training, test and validation sets using 10-fold cross validation, and projecting the edges of each fold onto the layers; for instance, if the edge $(v, u)$ was in the current training/test/validation split, we took it in the training/test/validation split of layer $l$ only if it appeared in layer $l$. The negative training/test/validation non-linked node-pairs were randomly sampled for each layer, in the same amount of the positive ones. First, we trained each individual expert separately, and then we trained MoE-ML-LP while keeping the experts frozen. To ensure a fair comparison, we used the same positive and negative node-pairs for all methods.

**Parameters setting in MoE-ML-LP.** We determined the optimal set of hyperparameters for the gating module $G(\cdot)$ by performing a grid-search over the configurations detailed in Appendix D.3. The selected configuration includes 2 hidden layers with a hidden size of 16, a learning rate of 1e-3, a weight decay of 1e-4, a dropout rate of 0.3, and a batch size of 128. For the training, we used early stopping with a patience of 5 epochs, based on the validation set loss.

The heuristics, resp. experts, involved in MoE-ML-LP correspond to the multilayer extension (based on Eq. (1)) of the baselines, resp. competing methods described next. For the definition of multilayer heuristics, we set $\alpha = 0.5$ as an informativeness trade-off between the target and remaining layers.

**Baselines and Competing Methods** MoE-ML-LP is compared with *(i)* multilayer extensions of traditional methods of link predictions and *(ii)* more sophisticated methods conceived for multilayer link prediction. We will refer to the former as *baselines* and to the latter as *competing methods*. The baselines include our defined multilayer extensions of Adamic-Adar, Common Neighbors, Jaccard, and Personalized Page Rank (cf. Appendix A), hereinafter denoted as *mAA*, *mCN*, *mJC*, and *mPPR*, respectively. The competing methods include BPHGNN [Fu *et al.*, 2023], GATNE [Cen *et al.*, 2019], HDMI [Jing *et al.*, 2021], and MAGMA [Coscia *et al.*, 2022]. We also introduce two *ensemble-based strategies*: a simple mean-based aggregation ($Ens_S$), and a weighted ensemble approach ($Ens_W$), which combine the predictions of individual methods (see Appendix D.1).

| Model | Cs-Aarhus | CKM | Lazega | Elegans | DkPol | ArXiv |
|---|---|---|---|---|---|---|
| mAA | 0.58 ± .13 | 0.33 ± .07 | 0.20 ± .07 | 0.18 ± .08 | 0.19 ± .03 | **0.88 ± .05** |
| mCN | 0.53 ± .12 | 0.28 ± .05 | 0.19 ± .07 | 0.15 ± .05 | 0.19 ± .04 | 0.66 ± .08 |
| mJC | 0.54 ± .11 | 0.29 ± .07 | 0.18 ± .05 | 0.07 ± .02 | 0.09 ± .02 | 0.79 ± .06 |
| mPPR | 0.54 ± .11 | 0.43 ± .13 | 0.17 ± .05 | 0.35 ± .09 | 0.04 ± .01 | OOT |
| BPHGNN | 0.40 ± .16 | 0.26 ± .09 | 0.18 ± .06 | 0.10 ± .02 | 0.10 ± .03 | 0.04 ± .00 |
| GATNE | 0.46 ± .14 | 0.36 ± .12 | 0.13 ± .04 | 0.17 ± .05 | 0.03 ± .01 | 0.55 ± .08 |
| HDMI | 0.48 ± .09 | 0.23 ± .06 | 0.20 ± .05 | 0.25 ± .07 | 0.06 ± .02 | 0.39 ± .10 |
| MAGMA | 0.36 ± .10 | 0.38 ± .13 | 0.15 ± .05 | 0.48 ± .12 | 0.19 ± .05 | OOT |
| $Ens_S$ | 0.53 ± .15 | 0.38 ± .10 | 0.19 ± .05 | 0.36 ± .12 | 0.18 ± .04 | 0.53 ± .09 |
| $Ens_W$ | 0.45 ± .15 | 0.38 ± .12 | 0.16 ± .06 | 0.23 ± .08 | 0.19 ± .04 | 0.07 ± .01 |
| **Ours** | **0.72 ± .15** | **0.60 ± .16** | **0.33 ± .08** | **0.73 ± .16** | **0.54 ± .20** | 0.85 ± .13 |

Table 1: MRR results on the test sets of real-world networks. Bold and underlined values indicate the best and second-best scores, respectively. OOT denotes Out-Of-Time (i.e., running time > 24h).

| Model | Cs-Aarhus | CKM | Lazega | Elegans | DkPol | ArXiv |
|---|---|---|---|---|---|---|
| mAA | 0.36 ± .18 | 0.16 ± .11 | 0.09 ± .09 | 0.09 ± .10 | 0.10 ± .05 | **0.84 ± .06** |
| mCN | 0.30 ± .19 | 0.18 ± .12 | 0.10 ± .09 | 0.07 ± .06 | 0.11 ± .06 | 0.61 ± .09 |
| mJC | 0.34 ± .17 | 0.11 ± .08 | 0.06 ± .05 | 0.01 ± .02 | 0.02 ± .02 | 0.74 ± .07 |
| mPPR | 0.26 ± .21 | 0.23 ± .18 | 0.04 ± .03 | 0.18 ± .10 | 0.00 ± .01 | OOT |
| BPHGNN | 0.22 ± .21 | 0.13 ± .10 | 0.08 ± .06 | 0.04 ± .02 | 0.03 ± .02 | 0.02 ± .01 |
| GATNE | 0.28 ± .20 | 0.19 ± .13 | 0.04 ± .04 | 0.06 ± .05 | 0.01 ± .01 | 0.40 ± .15 |
| HDMI | 0.34 ± .13 | 0.14 ± .09 | 0.09 ± .05 | 0.14 ± .08 | 0.02 ± .02 | 0.32 ± .12 |
| MAGMA | 0.14 ± .15 | 0.20 ± .16 | 0.05 ± .07 | 0.34 ± .18 | 0.12 ± .07 | OOT |
| $Ens_S$ | 0.31 ± .22 | 0.20 ± .14 | 0.07 ± .05 | 0.21 ± .15 | 0.11 ± .06 | 0.39 ± .11 |
| $Ens_W$ | 0.26 ± .21 | 0.21 ± .15 | 0.07 ± .07 | 0.14 ± .08 | 0.10 ± .06 | 0.03 ± .02 |
| **Ours** | **0.53 ± .29** | **0.44 ± .22** | **0.16 ± .07** | **0.64 ± .20** | **0.38 ± .28** | 0.75 ± .25 |

Table 2: Hits@1 results on the test sets. Bold and underlined values indicate the best and second-best scores, respectively.

## 6 Results

### 6.1 Comparative Evaluation

**Real-world networks.** Table 1 reports the MRR values on real-world datasets. MoE-ML-LP outperforms all methods on all networks, except for *ArXiv*, where it closely follows mAA. Overall, MoE-ML-LP achieves an average MRR improvement of +60% w.r.t. competing methods and baselines.

A closer examination reveals a number of remarks. On *Cs-Aarhus*, MoE-ML-LP achieves a +24% improvement in MRR compared to the second-best model (mAA), and a +50% improvement over the best expert. On *CKM*, MoE-ML-LP shows a +39.5% MRR increase over the second-best approach (mPPR), and a +58% gain over the best expert (MAGMA). On *Lazega*, MoE-ML-LP delivers an impressive +65% improvement in MRR compared to the second-best models (HDMI and mAA). This also holds on *Elegans*, where MoE-ML-LP shows a +52% MRR improvement over the best expert. Even better, on *DkPol*, MoE-ML-LP outperforms the second-best models by +184%, which is particularly significant as MAGMA. Remarkably, this demonstrates the MoE-ML-LP's ability to enhance the predictions of its single experts. Lastly, on *ArXiv*, while MoE-ML-LP achieves the second-best score with an MRR just 2% lower than mAA, it still achieves a +54.5% improvement in MRR compared to the best-performing single expert. This result is likely to be ascribed to the particularly high clustering coefficient and strong community structure of *ArXiv* (cf. Appendix B), where heuristic-based methods tend to excel.

Consistently with the MRR results, MoE-ML-LP signifi-cantly outperforms all other models, with an average +81.9% improvement in Hits@1, and +54.5%, resp. +40.6% increase in Hits@5, resp. Hits@10, compared to the other methods.

For the sake of brevity, here we discuss the Hits@1 results from Table 2, as it represents the most challenging scenario; we refer the reader to Appendix D.5 for details regarding Hits@5 and Hits@10 results. On *Cs-Aarhus*, MoE-ML-LP obtains a +47.2% improvement over the second-best model (mAA), and a +56% gain compared to the best-performing expert. On *CKM*, MoE-ML-LP achieves an impressive +91.3% increase over the second-best model and expert (mPPR), while on *Lazega*, it improves of 60% over both the second-best model (mAA) and expert (HDMI). Similarly, MoE-ML-LP outperforms the second-best model and individual expert (MAGMA) by +88.2% on *Elegans* and by an astonishing +216.7% on *DkPol*. Moreover, as already observed with the MRR results, MoE-ML-LP expectedly ranks second to mAA on *ArXiv*, while improving over the best-performing individual expert (GATNE) by a +87.5% increase.

**Synthetic networks.** Our MoE-ML-LP is confirmed to outperform all methods according to all criteria, with significant gains against the best of the other models: +15.8% MRR, +11.6% Hits@1, and +22.7% Hits@5 w.r.t. mJC, and +10.5% Hits@10 w.r.t. MAGMA, in terms of averages over the test sets of all synthetic networks—see Table 11, Appendix D.5. The heuristics also perform well, even overall better than the experts (with the exception of GATNE), likely due to the small-world property of the evaluated synthetic networks which provides a relatively easier benchmark for these heuristics. We report details on the time performance on these synthetic networks in Appendix E.

**Remarks on Ensemble strategies.** Considering results on both real and synthetic data, the two ensemble approaches fall significantly short compared to MoE-ML-LP and, in some cases, perform even worse than individual experts. This is expected, as the simple ensemble method merely averages the predictions, ignoring the different relevance of each expert, while the weighted ensemble learns and applies a single weighing strategy across all $(u, v, l)$ triples, failing to capture the expert-specific importance for each case. These results underscore the key advantage provided by MoE-ML-LP, which learns how to combine experts' predictions, leading to substantial performance improvements.

## 6.2 Sensitivity Analysis

**Sensitivity to heuristic removal.** We evaluated the impact of the heuristics used in the gating module of MoE-ML-LP (i.e., mAA, mPPR, mJC, and mCN). Figure 2 (top) shows the average MRR and Hits@10 results over all real-world networks for each combination of heuristics. Results for Hits@1 and Hits@5 are provided in Appendix F. It can be noted a clear decrease in performance as the number of involved heuristics narrows, with the best performance achieved by using the full set of heuristics. This demonstrates that MoE-ML-LP can effectively benefit from both local (such as mAA, mJC, and mCN) and global structural information (i.e., mPPR), leading to consistently better results (cf. Appendix F).

On the other hand, there is no evidence of a diminishing-returns trend, suggesting that adding more heuristics could in principle still lead to improvements. However, we conjecture that including more heuristics with similar characteristics (e.g., those relying on local topological structures like AA, JC, CN, or random-walk-based strategies like PPR) might reduce the benefit of enlarging the heuristic set. This can be explained by the overlapping nature of the information the heuristics capture. For instance, heuristics like AA, JC, and CN rely on shared neighborhoods, potentially influencing the gating module similarly. While a reasonable number of heuristics can provide complementary insights, adding too many similar ones may introduce redundancy, resulting in diminishing returns. This is because the gating module becomes less likely to benefit from the diversity of insights that is expected by different heuristics.

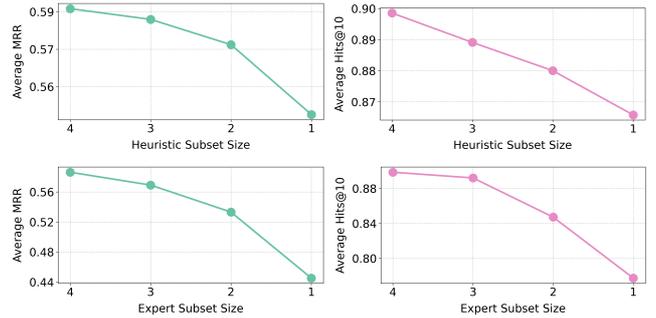

Figure 2: Average performance by varying the size of the set of heuristics (top) and the size of the set of experts (bottom).

**Sensitivity to expert removal.** We also assessed the impact of varying the number of experts in MoE-ML-LP by evaluating all possible subsets of the experts used in our study (i.e., BPHGNN, GATNE, HDMI, and MAGMA). Figure 2 (bottom) shows the average MRR and Hits@10 results over all real-world networks for each combination of experts. Results for Hits@1 and Hits@5 are provided in Appendix F. The figure shows that the average MRR decreases as the number of experts is reduced. In particular, halving the number of experts results in a ∼10% decrease in MRR, while using only a single expert leads to ∼24% reduction. Similar trends hold for the Hits@$k$ metrics, where halving the number of experts, resp., considering only one expert, leads to an average reduction of 8.5%, resp., 22.4%, by varying $k$ (cf. Appendix F). This underscores the importance of using all available experts, as Figure 3 shows many edges correctly predicted by individual experts but missed by others.

In addition, Figure 2 (bottom) also shows a diminishing-returns trend with an increasing number of experts. We ascribe this trend to the likelihood that additional models might rely on similar assumptions or strategies, thus reducing their informativeness. Moreover, as will be discussed in the next paragraph (cf. Figure 4), our gating module effectively activate the most relevant experts for each triple $(u, v, l)$, meaning that, even with more experts, only the most informative ones are prioritized, making the addition of redundant experts ineffective in terms of both resources and performance.

**Analysis of gating weights.** Figure 4 shows the gating weights (cf. Eq. (2)) averaged over the test set of each real-world network. Notably, all or most experts in MoE-ML-LP contribute to the final predictions, indicating that the *collapse problem* [Shazeer et al., 2017], commonly observed in MoE models (i.e., a single expert consistently favored, resulting in the under-utilization of others), does not occur in our approach: in fact, for any particular network we notice that different experts are activated to handle distinct node pairs across layers. One exception is BPHGNN, which is rarely activated, probably due to its lower ability to provide unique correct predictions w.r.t. the other experts, as shown in Figure 3. Notably, *ArXiv* is the only case where BPHGNN receives a high attention weight, as it is the second-best expert in terms of unique correct predictions (cf. Figure 3). Additionally, the highest improvement in MRR of MoE-ML-LP compared to the individual experts is observed on *DkPol*, where 3 out of 4 experts contribute a significant fraction of unique correct predictions (above 30%, cf. Figure 3). This highlights the effectiveness of the MoE approach in scenarios where it can leverage the strengths of multiple experts.

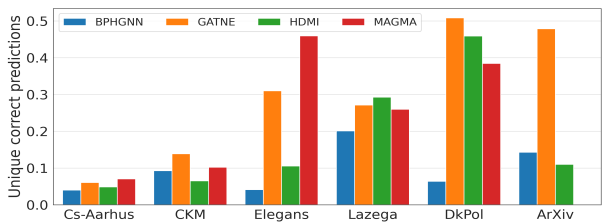

Figure 3: Fraction of unique correct predictions for each expert with respect to all other experts.

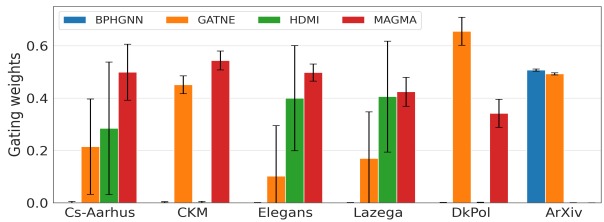

Figure 4: Averaged gating weights (across folds) with standard deviations of MoE-ML-LP for the experts.

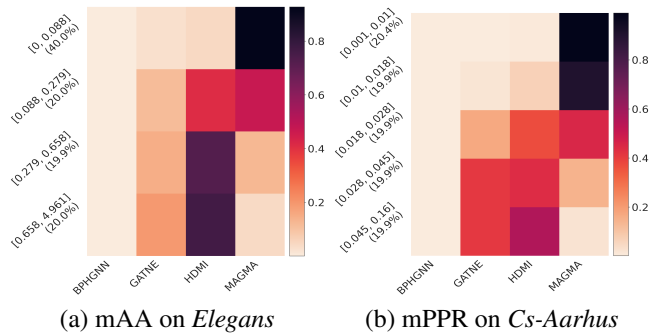

(a) mAA on *Elegans*  (b) mPPR on *Cs-Aarhus*

Figure 5: Distributions of gating weights assigned with experts according to various ranges of heuristic scores. Rows correspond to quantiles of heuristic scores, with round brackets reporting the percentage of node-pairs having linkage score in that quantile.

**Impact of heuristics on expert selection.** A further important aspect we investigated concerns the effect of the various heuristics, and their corresponding value ranges on the routing of the gating module toward individual experts. Results on selected examples are shown in Figure 5.

A first remark is about MAGMA consistently receiving the highest gating signal when heuristics' weights are in a low regime. This behavior is likely due to the association-rules-based nature of MAGMA, which might prove more beneficial when heuristics assign a low linkage probability based on structural properties. More specifically, since MAGMA searches for multilayer patterns of varying sizes to infer its own rules, we argue that MAGMA is capable of capturing more complex information about the subgraph containing a particular target link, compared to heuristics that focus solely on the endpoints associated to the target link.

Also, in accord with the previously discussed analysis of gating weights, we notice that BPHGNN tends to receive the least weight across all heuristic-dataset combinations.

Two main patterns on the heuristics' influence are observed. The first one involves specific heuristics, especially under high values, amplifying the gating weight toward particular experts, such as HDMI and GATNE in the case of mAA, and GATNE for mCN (not shown). By contrast, heuristics like mJC and mPPR result in a more evenly distributed gating weight, often balancing between the two most prominent experts for a given network.

Overall, different heuristics guide the gating module to adaptively weight experts based on the input network and triples. This confirms our earlier findings from the gating weight evaluation and reinforces the importance of diverse local and global heuristics for optimizing expert weights.

## 7 Conclusions

Link prediction in multilayer networks is challenging due to structural complexities within and across layers. Existing methods often overlook the rich information in these networks by focusing on limited aspects that are peculiar for some layers, hindering their ability to fully capture the broader and multilayered network characteristics. To address this, we propose MoE-ML-LP, the first Mixture-of-Experts (MoE) framework for multilayer link prediction. MoE-ML-LP uses traditional link prediction heuristics adapted to multilayer settings to combine the strengths of diverse experts, resulting in improved predictive performance. Extensive experiments on real-world and synthetic datasets show that MoE-ML-LP outperforms traditional baselines and advanced methods, achieving notable improvements in MRR and Hits@$k$. Its modular design allows for easy integration of new experts to further enhance performance. Future work will explore applying MoE-ML-LP to more complex settings, such as temporal or heterogeneous multilayer networks, to extend its predictive capabilities.

**Remarks on reproducibility.** We are committed to making all code publicly available upon acceptance.

# Heuristic-Informed Mixture of Experts for Link Prediction in Multilayer Networks
# TECHNICAL APPENDIX

## A Heuristic Methods for Link Prediction in Single-Layer Graphs

Here we present an overview the traditional single-layer heuristic algorithms we used as building blocks for the definition of our multilayer heuristic scores (see Eq. 1 in the main paper). Below, we will use $\Gamma_u$ to indicate the neighborhood of node $u$ in a single-layer graph.

**Common Neighbors (CN)** [Liben-Nowell and Kleinberg, 2003] estimates the likelihood of a connection between two nodes by counting the number of shared neighbors:

$$CN(u,v) = |\Gamma_u \cap \Gamma_v|. \qquad (6)$$

CN is widely applied in social networks for tasks like friend recommendations, as it has shown a correlation between the number of common neighbors and the likelihood of a link forming between two nodes [Newman, 2001].

**Jaccard similarity.** Akin to CN, the Jaccard index measures the overlap of neighborhoods between two nodes:

$$JC(u,v) = \frac{|\Gamma_u \cap \Gamma_v|}{|\Gamma_u \cup \Gamma_v|}. \qquad (7)$$

It normalizes the CN score by comparing the fraction of shared nodes to the overall neighbors, providing a probabilistic view of the commonality between the two nodes.

**Adamic-Adar** [Adamic and Adar, 2003] quantifies the similarity between two nodes by assigning a weighted importance to common neighbors:

$$AA(u,v) = \sum_{w \in \Gamma_u \cap \Gamma_v} \frac{1}{\log |\Gamma_w|}, \qquad (8)$$

where the degree of each common neighbor is logarithmically penalized, as the main assumption is that low-degree nodes carry more informative value, and are therefore given higher weight in the calculation.

**Personalized PageRank (PPR) score** [Brin and Page, 1998]. It can be used for link prediction by evaluating the probability that a random walk starting from one node reaches another. Given the transition matrix $\mathbf{P}$ and a personalization vector $\mathbf{e}_u$, which is a one-hot vector with a 1 at node $u$), the Personalized PageRank vector $\mathbf{r}_u$ is computed as:

$$\mathbf{r}_u = (1-\beta)\mathbf{e}_u + \beta \mathbf{P}^T \mathbf{r}_u \qquad (9)$$

where $\beta \in [0,1]$ is the probability of following an edge in the random walk. The PPR score $r_u(v)$ measures the likelihood of reaching node $v$ from $u$, and can be used directly as the link prediction score. To ensure symmetry, the final link prediction score between $u$ and $v$ can be:

$$PPR(u,v) = \frac{r_u(v) + r_v(u)}{2} \qquad (10)$$

This approach leverages node proximity and importance to predict potential links.

## B Data Description

Here we describe the real-world multilayer networks used in our evaluation, whose main structural characteristics are reported in Table 3.

**Cs-Aarhus** [Magnani et al., 2013] is a social network consisting of five types of (undirected) relationships among employees (nodes) in the Department of Computer Science at Aarhus University. Layers corresponds to online and offline relations: Co-authorship, Facebook, Leisure, Lunch, Work relations.

**CKM** [Coleman et al., 1957] is a network based on social interactions among physicians (nodes) when adopting new drugs. It consists of three directed layers capturing different types of interactions: advice-seeking, discussion, and friendship.

**Lazega** [Kraatz et al., 2003] is a directed social network among partners and associates in a corporate law partnership, where its three layers represent different forms of interaction: Advice, Friendship, and Co-workship relationship.

**Elegans** [Chen et al., 2006] is an undirected multiplex network representing the connectome of the Caenorhabditis elegant nematode. Entities are neurons and layers correspond to different types of synaptic junction: electric, chemical monadic, and polyadic.

**DkPol** (Dansk Politik) [Magnani et al., 2022] is a directed Twitter of Danish politicians running for the parliament, comprising three types of online interaction: Retweet, Reply, and Follow.

**ArXiv** [De Domenico et al., 2015a] is an undirected co-authorship network featuring 13 layers, each representing different arXiv categories.

## C Computational Time Complexity

Considering the computational complexity of our framework, we emphasize that the multilayer heuristics can be seamlessly pre-computed before training the gating module. In a similar manner, since the experts are frozen during the training phase of MoE-ML-LP, their training procedure can be regarded as a preprocessing for the training of MoE-ML-LP.

The computational cost of our framework during the training stage is dominated by the gater, which is implemented as a two-layer MLP with a hidden dimension of $d$. Specifically, the first layer of the MLP involves multiplying the heuristics-generated input vector of dimension $k$ by a weight matrix of size $d \times k$, resulting in a computational cost of $\mathcal{O}(kd)$, where $k$ is the number of multilayer heuristics. The second layer involves multiplying the hidden activations of dimension $d$ by a weight matrix of size $n \times d$, with a computational cost of $\mathcal{O}(nd)$. The softmax activation function applied to the $n$ logits costs $\mathcal{O}(n)$, which is negligible compared to the previous terms. Consequently, to process an edge, the total computational complexity of the forward pass is $\mathcal{O}(kd + nd)$. Additionally, aggregating the experts' predictions involves computing a weighted sum over $n$ experts, which incurs a computational cost of $\mathcal{O}(n)$. Consequently, the total computational complexity of the forward pass is $\mathcal{O}(kd + nd)$, which is $\mathcal{O}(|E_\mathcal{L}|d(k + n))$ over all the intra-layer edges. Since $k$ and $n$ are typically small constants, the computational complexity is approximately linear with respect to both the hidden dimension $d$ and the number of edges $|E_\mathcal{L}|$. This implies that the model scales efficiently as the size of the data increases, maintaining manageable computational demands. We also emphasize that each heuristic and expert's predictions can be executed in parallel.

| Network | $|\mathcal{V}|$ | $|V_\mathcal{L}|$ | $|E_\mathcal{L}|$ | $\ell$ | Avg. Degree | Avg. Path Len. | Clust. Coeff. | Density | Diameter | Modularity | Num. Communities |
|---|---|---|---|---|---|---|---|---|---|---|---|
| Cs-Aarhus | 61 | 224 | 620 | 5 | 1.680 | 1.667 | 0.429 | 0.073 | 8 | 0.757 | 8 |
| | | | | | 7.750 | 1.956 | 0.481 | 0.250 | 4 | 0.332 | 4 |
| | | | | | 3.745 | 3.123 | 0.343 | 0.081 | 8 | 0.571 | 6 |
| | | | | | 6.433 | 3.189 | 0.569 | 0.109 | 7 | 0.654 | 5 |
| | | | | | 6.467 | 2.390 | 0.339 | 0.110 | 4 | 0.452 | 4 |
| CKM | 246 | 674 | 1551 | 3 | 2.233 | 3.481 | 0.212 | 0.010 | 6 | 0.722 | 8 |
| | | | | | 2.446 | 4.504 | 0.211 | 0.011 | 14 | 0.740 | 8 |
| | | | | | 2.219 | 3.669 | 0.241 | 0.010 | 10 | 0.759 | 8 |
| Lazega | 71 | 211 | 2571 | 3 | 12.563 | 2.243 | 0.365 | 0.179 | 6 | 0.281 | 3 |
| | | | | | 8.333 | 2.505 | 0.347 | 0.123 | 7 | 0.369 | 4 |
| | | | | | 15.549 | 1.886 | 0.341 | 0.222 | 4 | 0.301 | 3 |
| Elegans | 279 | 791 | 5860 | 3 | 4.063 | 4.523 | 0.128 | 0.016 | 12 | 0.638 | 11 |
| | | | | | 6.304 | 3.436 | 0.115 | 0.024 | 9 | 0.483 | 9 |
| | | | | | 11.486 | 2.749 | 0.207 | 0.041 | 7 | 0,432 | 6 |
| DkPol | 490 | 839 | 20,198 | 3 | 4.321 | 3.965 | 0.176 | 0.020 | 9 | 0.663 | 8 |
| | | | | | 2.321 | 4.404 | 0.011 | 0.017 | 11 | 0.636 | 16 |
| | | | | | 79.922 | 1.920 | 0.520 | 0.163 | 4 | 0.183 | 6 |
| ArXiv | 14,489 | 26,796 | 59,026 | 13 | $3.899 \pm 0.945$ | $5.598 \pm 2.618$ | $0.650 \pm 0.180$ | $0.03 \pm 0.002$ | $14.538 \pm 7.523$ | $0.942 \pm 0.04$ | $295.693 \pm 154.621$ |

Table 3: Main structural characteristics of the real-world networks. Modularity and Num. Communities denote values obtained with the Louvain algorithm, with resolution equal to 1.

## D Details on Experimental Methodology

### D.1 Additional baselines

Following [Ma *et al.*, 2024], we devised two straightforward aggregation strategies to serve as additional baselines in our experimental setup.

Given a set of $n$ experts $\{E^{(1)}, \ldots, E^{(n)}\}$ for the multi-layer link prediction task, the first strategy simply aggregates the predictions from the individual trained experts by taking their mean, which we refer to as $Ens_S$. This can be expressed as:

$$\hat{y}_{uvl} = \frac{1}{n}\sum_{i=1}^{n} E_{uvl}^{(i)} \qquad (11)$$

The second strategy, dubbed as $Ens_W$, involves learning a weight vector $\mathbf{w} = [w_1, w_2, \ldots, w_n]$ using a Multi-Layer Perceptron (MLP) to combine the predictions of each expert in a weighted manner. The final score is computed as:

$$\hat{y}_{uvl} = \sigma\left(\sum_{i=1}^{n} \mathbf{w}_i \cdot E_{uvl}^{(i)}\right) \qquad (12)$$

where $\sigma$ is the sigmoid activation function.

It is important to note that this method differs from MoE-ML-LP, as the learned weights remain fixed across all $(u, v, l)$ triples, thus acting as a simpler MoE baseline.

### D.2 Implementation details

We implemented our method using PyTorch[1] library, while for the implementation of the heuristic algorithms we used the NetworkX library.[2] For GATNE [Cen *et al.*, 2019],[3] BPHGNN [Fu *et al.*, 2023],[4] HDMI[Jing *et al.*, 2021],[5] and MAGMA [Coscia *et al.*, 2022],[6] we used their publicly available source code.

Our experiments were conducted using a double 56-core Intel(R) Xeon(R) Gold 6258R CPU, equipped with 256GB RAM and two NVIDIA GeForce RTX3090s with 24GB memory each, OS Ubuntu Linux 22.04 LTS.

### D.3 Hyper-parameters

**Experts hyper-parameters.** For training all the experts, we used the default hyper-parameters provided in their source code and/or in their corresponding paper. Specifically, for GATNE, BPHGNN and HDMI we used the default hyper-parameters from their source code for each dataset. Also, BPHGNN requires the computation of basic behavior patterns [Fu *et al.*, 2023], which are in exponential number w.r.t. the number of layers. We computed all the basic behavior patterns for all datasets, with the exception of ArXiv due to its high number of layers (13). Specifically, we used BPHGNN on ArXiv with basic behavior patterns of length 2. For HDMI and BPHGNN, we predicted links by computing the dot product between the node representations learned at each layer. For MAGMA, we used the same hyperparameters as in their paper [Coscia *et al.*, 2022] for the datasets common to both our work and theirs (Cs-Aarhus, CKM, Elegans), specifically setting the maximum pattern size to 4, confidence to 0, and support to 15, 20, and 75, respectively. For the ArXiv dataset, we set the maximum pattern size to 4, confidence to 0, and support to 5. For Lazega and DkPol, we reduced the maximum pattern size due to time constraints, using a maximum pattern size of 3, confidence of 0, and support of 75. We used the identity matrix as node attributes for all the experts that require them.

---

[1] https://pytorch.org/
[2] https://networkx.org/
[3] https://github.com/THUDM/GATNE
[4] https://github.com/FuChF/BPHGNN-23
[5] https://github.com/baoyujing/HDMI
[6] https://www.michelecoscia.com/?page_id=1857

For the efficiency analysis, we selected the same hyperparameter settings as in our main experiments. For MAGMA we set confidence, support, and maximum pattern size to 0, 15, and 3, respectively. To ensure a fair comparison, we used a fixed number of epochs for all methods, i.e., 100, without applying the early stopping technique where it is included in the default configuration.

**Grid search over MoE-ML-LP parameters.** During the training of MoE-ML-LP, we experimented with various configurations to find the optimal set of hyperparameters for our gating module $G(\cdot)$. We tested different combinations, including a number of hidden layers between 1 and 2, the hidden size in $\{16, 32, 64\}$, the learning rate between in $\{1e\text{-}3, 1e\text{-}4\}$, the dropout rate in $\{0.1, 0.3\}$, and the batch size in $\{32, 64, 128\}$. To determine the ideal number of epochs, we employed early stopping with patience of 5 epochs based on the validation loss. Finally, we ranged $\alpha \in [0.1, 0.9]$ with a 0.2 step in the multilayer heuristics calculation.

### D.4 Synthetic networks generation and results

| $|\mathcal{V}|$ | $|V_\mathcal{L}|$ | $|E_\mathcal{L}|$ | Avg. Degree | Avg. Path Len. | Clust. Coeff. |
|---|---|---|---|---|---|
| 500 | 1500 | 1400 | 6.667 | 12.895 | 0.321 |
| 1000 | 3000 | 14000 | 9.333 | 5.817 | 0.457 |
| 2000 | 6000 | 44000 | 14.667 | 4.245 | 0.503 |
| 4000 | 12000 | 148000 | 24.667 | 3.650 | 0.523 |

Table 4: Main structural characteristics of the synthetic networks.

For the sake of completeness in our experimental setup, we also generated four synthetic networks with 3 layers and a different number of entities: 500, 1000, 2000 and 4000, using the Watts-Strogatz model with rewiring probability equal to 0.1. Each layer was generated with a different realization of Watts-Strogatz, and using the 0.5% of the number of entities as average degree for obtaining similar networks to the real ones. Table 4 reports the basic statistics of the synthetically generated networks.

### D.5 Additional results on real-world networks

**Unbiased sampling strategy.** In our main experiments, we randomly sampled positive and negative node pairs following the standard practice of several link prediction studies, e.g., [Zhang and Chen, 2018; Yun *et al.*, 2021]. Moreover, to ensure a fair comparison, we applied the same sets of positive/negative pairs to both the competing methods and our own approach. However, to further evaluate the robustness of our framework, we evaluated its performance by sampling negative examples at each layer according to the approach described in [Aiyappa *et al.*, 2024], which is designed to mitigate the mismatch in degree distributions between positive and negative edges. Table 5 presents the results on the Cs-Aarhus network, which comprises five layers, each featuring a distinct average degree (cf. Table 3).

| Model | MRR | Hits@1 |
|---|---|---|
| GATNE | 0.13 ± .10 | 0.05 ± .08 |
| MAGMA | 0.13 ± .07 | 0.02 ± .04 |
| HDMI | 0.17 ± .10 | 0.07 ± .08 |
| BPHGNN | 0.14 ± .09 | 0.06 ± .10 |
| **ML-MoE-LP** | **0.19 ± .12** | **0.12 ± .12** |

Table 5: Results on Cs-Aarhus with the degree unbiased negative sampling approach.

| Model | MRR | Hits@1 | Hits@10 |
|---|---|---|---|
| Neo-GNN | 0.45 ± 0.18 | 0.29 ± 0.22 | 0.84 ± 0.11 |
| SEAL | 0.29 ± 0.09 | 0.16 ± 0.09 | 0.57 ± 0.18 |
| NCN | 0.56 ± 0.12 | 0.41 ± 0.17 | 0.91 ± 0.06 |
| NCNCN | 0.56 ± 0.15 | 0.41 ± 0.22 | 0.89 ± 0.07 |
| MoE-LP | 0.69 ± 0.17 | 0.49 ± 0.30 | 0.99 ± 0.03 |
| **ML-MoE-LP** | **0.72 ± 0.15** | **0.53 ± 0.29** | **0.99 ± 0.02** |

Table 6: Results on Cs-Aarhus for MoE-ML-LP and single-layer methods.

| Model | MRR | Hits@1 | Hits@10 |
|---|---|---|---|
| Neo-GNN | 0.34 ± 0.06 | 0.16 ± 0.10 | 0.72 ± 0.08 |
| SEAL | 0.24 ± 0.07 | 0.08 ± 0.08 | 0.63 ± 0.09 |
| NCN | 0.38 ± 0.11 | 0.21 ± 0.16 | 0.71 ± 0.06 |
| NCNCN | 0.39 ± 0.09 | 0.22 ± 0.12 | 0.70 ± 0.09 |
| MoE-LP | 0.56 ± 0.18 | 0.39 ± 0.23 | 0.89 ± 0.07 |
| **ML-MoE-LP** | **0.60 ± 0.16** | **0.44 ± 0.22** | **0.98 ± 0.03** |

Table 7: Results on CKM for MoE-ML-LP and single-layer methods.

| Model | MRR | Hits@1 | Hits@10 |
|---|---|---|---|
| Neo-GNN | 0.18 ± 0.05 | 0.09 ± 0.05 | 0.38 ± 0.09 |
| SEAL | 0.17 ± 0.05 | 0.06 ± 0.06 | 0.38 ± 0.05 |
| NCN | 0.24 ± 0.06 | 0.08 ± 0.07 | 0.55 ± 0.08 |
| NCNCN | 0.23 ± 0.06 | 0.08 ± 0.08 | 0.54 ± 0.08 |
| MoE-LP | 0.32 ± 0.10 | 0.15 ± 0.12 | 0.68 ± 0.04 |
| **ML-MoE-LP** | **0.33 ± 0.08** | **0.16 ± 0.07** | **0.70 ± 0.13** |

Table 8: Results on Lazega for MoE-ML-LP and single-layer methods.

This new sampling strategy is more challenging for any link prediction algorithm [Aiyappa *et al.*, 2024], as testified by the reduced overall scores concerning all baselines, which are also found on our approach. Furthermore, we observe that this approach also changes the ranking of the baseline methods. Nonetheless, our proposed MoE-ML-LP still remains the best performer, demonstrating robustness against this un-biased sampling strategy, achieving a +11.76% improvement in MRR, and +71.43% in HITS@1, with respect to the best expert.

| Model    | MRR          | Hits@1       |
|----------|--------------|--------------|
| GATNE    | 0.03 ± .00   | 0.02 ± .00   |
| HDMI     | 0.06 ± .01   | 0.04 ± .02   |
| BPHGNN   | 0.04 ± .02   | 0.02 ± .02   |
| **ML-MoE-LP** | **0.08 ± .00** | **0.06 ± .00** |

Table 9: Results on Homo Multiplex GPI Network for MoE-ML-LP and competing methods.

**Other networks.** We also tested MoE-ML-LP on the **Homo Multiplex GPI Network** [Stark *et al.*, 2006; De Domenico *et al.*, 2015b] consisting of 7 layers, more than 18K nodes, and 170K edges, as reported in Table 9. For this experiment, we do not report the results for the competitor method MAGMA, nor do we use it as an expert in our approach, as it ran out-of-time (i.e., running time above 24h).

It is worth noting that while our approach is designed to work seamlessly across networks of any size, most existing multi-layer link prediction experts are limited to handling small networks due to scalability constraints (e.g., MAGMA already goes OOT with reasonable-sized networks. Consequently, experimenting with very large networks (e.g., millions of nodes and edges) is restricted by their limitations, not the capabilities of our approach.

**Comparison with extended single-layer competing methods.** We also compared MoE-ML-LP with the current state-of-the-art single-layer approaches, namely Neo-GNN [Yun *et al.*, 2021], SEAL [Zhang and Chen, 2018], NCN/NCNCN [Wang *et al.*, 2024] and MoE-LP [Ma *et al.*, 2024]. For extending them to the multi-layer setting, we opted for feeding them each layer independently, and we then aggregated their predictions into stronger ones. We emphasize that this is a particularly favorable setting for these competing methods, as it allows them to focus on each layer independently (whereas our approach accounts for all of them simultaneously), thus maximizing their performance.

Nonetheless, as reported in Tables 6-8, our proposed approach still remains the best-performing one. Specifically, for the AUCS dataset, we achieved +28.6% in MRR compared to the best single expert (NCN/NCNC) and +4.4% compared to the multi-layer extension of Link-MoE, +29.3% in HITS@1 compared to the best single expert (NCN/NCNC), and +8.2% compared to the multi-layer extension of Link-MoE. For the CKM dataset, we achieved +54% in MRR compared to the best single expert (NCNC), and +7.2% compared to the multi-layer extension of Link-MoE, +100% in HITS@1 compared to the best single expert (NCNC), and +12.8% compared to the multi-layer extension of Link-MoE. Finally, for the Lazega dataset, we achieved +37.5% in MRR compared to the best single expert (NCN), and +3.13% compared to the multi-layer extension of Link-MoE, +78% in HITS@1 compared to the best single expert (Neo-GNN), and +7% compared to the multi-layer extension of Link-MoE.

Overall, we report an average improvement in performance of +40% in MRR and +69.1% in HITS@1 over the best multi-layer extension of SOTA single-layer experts, and +5% in MRR and 9.3% in HITS@1 over the multi-layer extension of Link-MoE.

| Model | Cs-Aarhus | CKM | Lazega | Elegans | DkPol | ArXiv |
|-------|-----------|-----|--------|---------|-------|-------|
| mAA | 0.36 ± .18 | 0.16 ± .11 | 0.09 ± .09 | 0.09 ± .10 | 0.10 ± .05 | **0.84 ± .06** |
|     | 0.85 ± .06 | 0.55 ± .08 | 0.30 ± .09 | 0.23 ± .08 | 0.26 ± .04 | 0.93 ± .04 |
|     | 0.96 ± .04 | 0.71 ± .05 | 0.39 ± .09 | 0.39 ± .05 | 0.36 ± .03 | 0.97 ± .01 |
| mCN | 0.30 ± .19 | 0.18 ± .12 | 0.10 ± .09 | 0.07 ± .06 | 0.11 ± .06 | 0.61 ± .09 |
|     | 0.82 ± .09 | 0.58 ± .10 | 0.27 ± .09 | 0.24 ± .08 | 0.26 ± .05 | 0.77 ± .06 |
|     | 0.95 ± .04 | 0.71 ± .10 | 0.38 ± .10 | 0.34 ± .06 | 0.35 ± .03 | 0.90 ± .05 |
| mJC | 0.34 ± .17 | 0.11 ± .08 | 0.06 ± .05 | 0.01 ± .02 | 0.02 ± .02 | 0.74 ± .07 |
|     | 0.79 ± .10 | 0.51 ± .09 | 0.28 ± .08 | 0.07 ± .03 | 0.15 ± .04 | 0.90 ± .05 |
|     | 0.94 ± .05 | 0.70 ± .06 | 0.45 ± .08 | 0.14 ± .06 | 0.25 ± .04 | 0.95 ± .02 |
| mPPR | 0.26 ± .21 | 0.23 ± .18 | 0.04 ± .03 | 0.18 ± .10 | 0.00 ± .01 | OOT |
|      | 0.87 ± .04 | 0.65 ± .10 | 0.28 ± .10 | 0.56 ± .13 | 0.04 ± .02 | OOT |
|      | 0.95 ± .03 | 0.81 ± .08 | 0.47 ± .08 | 0.80 ± .10 | 0.11 ± .03 | OOT |
| BPHGNN | 0.22 ± .21 | 0.13 ± .10 | 0.08 ± .06 | 0.04 ± .02 | 0.03 ± .02 | 0.02 ± .01 |
|        | 0.61 ± .21 | 0.38 ± .14 | 0.24 ± .11 | 0.13 ± .03 | 0.14 ± .05 | 0.05 ± .02 |
|        | 0.81 ± .13 | 0.58 ± .08 | 0.38 ± .12 | 0.22 ± .06 | 0.26 ± .05 | 0.08 ± .01 |
| GATNE | 0.28 ± .20 | 0.19 ± .13 | 0.04 ± .04 | 0.06 ± .05 | 0.01 ± .01 | 0.40 ± .15 |
|       | 0.66 ± .11 | 0.57 ± .13 | 0.17 ± .06 | 0.27 ± .07 | 0.04 ± .02 | 0.77 ± .06 |
|       | 0.87 ± .08 | 0.79 ± .07 | 0.31 ± .06 | 0.38 ± .05 | 0.07 ± .02 | 0.87 ± .03 |
| HDMI | 0.34 ± .13 | 0.14 ± .09 | 0.09 ± .05 | 0.14 ± .08 | 0.02 ± .02 | 0.32 ± .12 |
|      | 0.65 ± .10 | 0.29 ± .06 | 0.31 ± .07 | 0.35 ± .10 | 0.08 ± .03 | 0.47 ± .10 |
|      | 0.76 ± .10 | 0.37 ± .05 | 0.45 ± .06 | 0.52 ± .07 | 0.14 ± .03 | 0.56 ± .08 |
| MAGMA | 0.14 ± .15 | 0.20 ± .16 | 0.05 ± .07 | 0.34 ± .18 | 0.12 ± .07 | OOT |
|       | 0.64 ± .13 | 0.61 ± .15 | 0.23 ± .06 | 0.66 ± .09 | 0.27 ± .06 | OOT |
|       | 0.89 ± .07 | 0.79 ± .06 | 0.37 ± .09 | 0.77 ± .08 | 0.36 ± .04 | OOT |
| $Ens_S$ | 0.31 ± .22 | 0.20 ± .14 | 0.07 ± .05 | 0.21 ± .15 | 0.11 ± .06 | 0.39 ± .11 |
|         | 0.76 ± .09 | 0.57 ± .12 | 0.30 ± .10 | 0.50 ± .12 | 0.25 ± .06 | 0.75 ± .06 |
|         | 0.91 ± .04 | 0.77 ± .08 | 0.44 ± .10 | 0.65 ± .09 | 0.35 ± .04 | 0.84 ± .05 |
| $Ens_W$ | 0.26 ± .21 | 0.21 ± .15 | 0.07 ± .07 | 0.14 ± .08 | 0.10 ± .06 | 0.03 ± .02 |
|         | 0.62 ± .09 | 0.54 ± .12 | 0.24 ± .07 | 0.43 ± .12 | 0.26 ± .05 | 0.10 ± .05 |
|         | 0.77 ± .10 | 0.77 ± .08 | 0.35 ± .09 | 0.42 ± .08 | 0.33 ± .02 | 0.13 ± .05 |
| **Ours** | **0.53 ± .29** | **0.44 ± .22** | **0.16 ± .07** | **0.64 ± .20** | **0.38 ± .28** | 0.75 ± .25 |
|          | **0.97 ± .06** | **0.83 ± .15** | **0.51 ± .15** | **0.85 ± .13** | **0.78 ± .11** | **0.98 ± .01** |
|          | **0.99 ± .02** | **0.98 ± .03** | **0.70 ± .13** | **0.91 ± .10** | **0.92 ± .09** | **0.98 ± .01** |

Table 10: Hits@k results on the test sets. Rows are grouped to show Hits@1, Hits@5, and Hits@10, respectively, for each method. Bold and underlined values indicate the best and second-best scores, respectively. OOT denotes Out-Of-Time (i.e., running time > 24h).

| Model | MRR | Hits@1 | Hits@5 | Hits@10 |
|-------|-----|--------|--------|---------|
| mAA | 0.46 ± .19 | 0.28 ± .26 | 0.69 ± .14 | 0.81 ± .07 |
| mCN | 0.41 ± .16 | 0.24 ± .21 | 0.69 ± .14 | 0.83 ± .06 |
| mJC | 0.57 ± .21 | 0.43 ± .26 | 0.75 ± .12 | 0.83 ± .06 |
| mPPR | 0.31 ± .19 | 0.16 ± .15 | 0.52 ± .33 | 0.72 ± .20 |
| BPHGNN | 0.06 ± .06 | 0.01 ± .01 | 0.07 ± .09 | 0.15 ± .15 |
| GATNE | 0.49 ± .19 | 0.34 ± .21 | 0.69 ± .18 | 0.82 ± .10 |
| HDMI | 0.27 ± .10 | 0.18 ± .10 | 0.37 ± .17 | 0.47 ± .15 |
| MAGMA | 0.26 ± .14 | 0.00 ± .00 | 0.43 ± .38 | 0.86 ± .05 |
| $Ens_S$ | 0.32 ± .19 | 0.10 ± .13 | 0.61 ± .26 | 0.82 ± .07 |
| $Ens_W$ | 0.20 ± .14 | 0.01 ± .00 | 0.35 ± .34 | 0.51 ± .28 |
| **Ours** | **0.66 ± .22** | **0.48 ± .33** | **0.92 ± .06** | **0.95 ± .04** |

Table 11: MRR and Hits@k results on the test sets of synthetic networks. Bold and underlined values indicate the best and second-best scores, respectively.

## E Efficiency Analysis

Table 12 reports the training times (in minutes) for each expert, along with MoE-ML-LP. Since MAGMA offers a CPU-only implementation, we ensure a fair comparison by reporting CPU training times for all other approaches. For GPU-compatible models, we observed faster training times that fol-low similar trends (results not shown).

| Model | $|\mathcal{V}| = 500$ | $|\mathcal{V}| = 1000$ | $|\mathcal{V}| = 2000$ | $|\mathcal{V}| = 4000$ |
|---|---|---|---|---|
| GATNE | 314.187 | 608.206 | > 12h | > 12h |
| MAGMA | 0.037 | 0.165 | 5.624 | 60.298 |
| HDMI | 0.163 | 0.474 | 0.779 | 1.474 |
| BPHGNN | 2.040 | 6.274 | 27.892 | OOM |
| Ours | 0.159 | 0.851 | 4.063 | 22.276 |

Table 12: Training times (in minutes) of the experts and MoE-ML-LP across different synthetically generated datasets by varying the number of nodes. OOM denotes unmanageable memory requirements.

MoE-ML-LP demonstrates consistent efficiency across all scenarios, maintaining scalability as graph sizes increase. This represents an additional strength for MoE-ML-LP. Indeed, although MoE-ML-LP is trained and applied after training experts, potentially adding computational overhead, the additional time required is minimal compared to the time needed to train most experts, deeming it particularly efficient. Notably, this modest overhead leads to significantly better performance, reinforcing both the scalability and efficiency of MoE-ML-LP.

## F  Additional Results on Sensitivity Analysis

Figure 6 shows the average Hits@1 and Hits@5 results over all real-world networks for each combination of heuristics (top) and experts (bottom). As with MRR and Hits@10 in the main paper (cf. Figure 2), performance in terms of Hits@1 and Hits@5 decreases when fewer heuristics are used, with the best results achieved by the full set, and no diminishing-returns trend, especially for Hits@5. Similarly, reducing the number of experts leads to performance drops in Hits@1 and Hits@5, with a diminishing-returns trend as the number of experts increases.

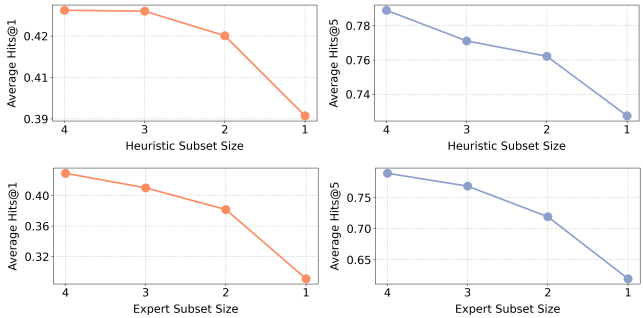

Figure 6: Average performance by varying the size of the set of heuristics (top) and the size of the set of experts (bottom).